%% file: main.tex
\documentclass[11pt]{article}

\usepackage[a4paper,margin=1in]{geometry}
\usepackage{times}
\usepackage{natbib}

\input{math_commands.tex}

\usepackage{multirow}
\usepackage{xcolor}
\usepackage{booktabs}
\usepackage{hyperref}
\usepackage{array}
\usepackage{url}
\usepackage{subcaption}   
\usepackage{graphicx}
\graphicspath{{fig/}}
\usepackage{amsmath}
\usepackage{wrapfig}
\usepackage{lipsum} 
\usepackage{amssymb} 
\usepackage{stmaryrd}

\title{ITC-MoE: Importance-guided Token-aware Compression for MoE Diffusion Language Models}

\author{
Lianjun Liu$^{1}$,
Shipeng Li$^{1}$,
You Huang$^{2}$,
Weiqi Yan$^{2}$,
Mingte Qiu$^{1}$,
Huazhong Liu$^{1,3}$,\\
Xiaofeng Zhu$^{1}$,
Yunshan Zhong$^{1}$\thanks{Corresponding author: \texttt{yszhong01@gmail.com}}\\[0.5em]
$^{1}$Hainan University, Haikou, China\\
$^{2}$Xiamen University, Xiamen, China\\
$^{3}$Zhejiang Normal University, Jinhua, China
}

\date{}
\begin{document}

\maketitle

\begin{abstract}
Mixture-of-Experts (MoE) Diffusion Language Models (DLMs) offer flexible parallel decoding and increased model capacity, but their large number of expert parameters incurs substantial computation and storage costs. 
Existing low-rank MoE compression methods largely rely on static factorization and fixed rank allocation, which overlook the distinctive properties of MoE DLMs. Specifically, we identify two properties: \emph{cross-mode non-uniform redundancy}, where parameter redundancy and sensitivity to rank truncation vary across the input, output, and expert modes, 
and \emph{token-wise utilization variation}, where hot and cold tokens exhibit distinct spectral characteristics and expert activation patterns. 
To address these challenges, we propose \textbf{ITC-MoE}, an Importance-guided Token-aware Compression framework for MoE DLMs. 
ITC-MoE consists of two complementary components. First, \textbf{Importance-guided Adaptive Tucker Compression (IATC)} incorporates activation and gradient importance into expert weight transformation, jointly factorizes expert weights across multiple modes, and adaptively allocates ranks under a fixed parameter budget.
Second, \textbf{Token-aware Compensation and Routing (TCR)} applies lightweight low-rank compensation to compression-sensitive hot tokens and restricts the candidate expert set for cold tokens with concentrated routing patterns. 
By jointly adapting compression capacity and inference execution to both parameter redundancy and token-wise variation, ITC-MoE substantially reduces the computation and storage costs of MoE DLMs while preserving their generation quality.
For example, on SDAR-30B-A3B-Chat-b32, ITC-MoE maintains an accuracy of 96.33\% on MultiArith under a 30\% compression budget, while achieving up to a $7.22\times$ end-to-end speedup. The code is publicly available at \url{https://github.com/lianjunl13-sudo/ITC-MoE}.

\end{abstract}

\input{paper/introduct}

\input{paper/related_work}

\input{paper/method}

\input{paper/exp}

\section{Limitations and Future Work}
Despite achieving a superior balance between speedup and model capacity, ITC-MoE still exhibits a performance gap relative to the full model, which can be alleviated by stronger compensation mechanisms. In addition, although candidate-restricted routing reduces the expert-selection complexity for cold tokens from \(O(E)\) to \(O(M)\), further hardware-aware implementations that fully exploit this optimization are still required to translate this algorithmic advantage more fully into end-to-end inference speedup.

\section{Conclusion}

In this paper, we identify two key properties of Mixture-of-Experts (MoE) Diffusion Language Models (DLMs): cross-mode non-uniform redundancy and token-wise utilization variation. The former stems from heterogeneous redundancy and truncation sensitivity across expert, input-feature, and output-feature modes, while the latter is reflected in the distinct spectral characteristics and expert activation patterns of hot and cold tokens.
Motivated by these observations, we propose Importance-guided Token-aware Compression (ITC-MoE), comprising importance-guided Adaptive Tucker Compression (IATC) and token-aware Compensation and Routing (TCR).
IATC performs importance-guided Tucker factorization and adaptively allocates ranks across modes under a fixed parameter budget. TCR mitigates information loss via lightweight low-rank compensation for compression-sensitive hot tokens, while applying candidate-restricted routing to cold tokens with concentrated expert activation patterns. Experiments across diverse DLMs and datasets confirm that, compared with existing methods, ITC-MoE achieves superior performance and efficiency.

\subsection*{Acknowledgments}
 
This work was supported by the National Natural Science Foundation of China (No. 62606133).

\subsection*{AI use statement}

In this work, we used generative AI tools to assist with language polishing, code development, and debugging. Generative AI tools were not used to generate experimental results or serve as a source of unverified research conclusions or literature evidence.
All AI-assisted content has been manually checked and verified by the authors. The authors take full responsibility for the final content of this paper, including all text, code, claims, results, and other related research materials.

\subsection*{Reproducibility statement}

To facilitate reproducibility, we have included source code and necessary experimental configurations in the supplementary material to facilitate reproduction of the main results reported in this work. The reproducibility of our paper is also supported by providing complete descriptions of the proposed method, implementation details, and experimental settings in the experimental section and appendix. Additional ablation studies and supplementary results are also provided in the appendix. 

\bibliographystyle{plainnat}
\bibliography{iclr2027_conference}

\appendix
\section{Appendix}

\subsection{Centered Kernel Alignment}
\label{sec:app-cka}

To measure the similarity of experts within a MoE layer, we employ Centered Kernel Alignment (CKA)~\citep{CKA}, which captures the geometric structure of output features. 
For any two experts $\mathbf{E}_i$ and $\mathbf{E}_j$ located in the same MoE layer, we first collect their output feature matrices $\mathbf{X}_{i}, \mathbf{X}_{j} \in \mathbb{R}^{B \times d}$ using the same set of $B$ calibration samples, where $d$ denotes the output dimension. We then center these features by subtracting the column-wise means:
\begin{equation}
\widetilde{\mathbf{X}}_{i}
=
\mathbf{J}_B\mathbf{X}_i,
\qquad
\widetilde{\mathbf{X}}_{j}
=
\mathbf{J}_B\mathbf{X}_{j},
\qquad
\mathbf{J}_B
=
\mathbf{I}_B-\frac{1}{B}\mathbf{1}_B\mathbf{1}_B^{\top},
\label{eq:center}
\end{equation}
where $\mathbf{J}_B$ is the centering matrix, $\mathbf{I}_B$ is the $B$-dimensional identity matrix, and $\mathbf{1}_B$ is the $B$-dimensional all-ones vector. Finally, the linear CKA between the two experts is defined as:
\begin{equation}
\operatorname{CKA}(\mathbf{X}_i,\mathbf{X}_{j})
=
\frac{
\left\|\widetilde{\mathbf{X}}_i^{\top}\widetilde{\mathbf{X}}_{j}\right\|_F^2
}{
\left\|\widetilde{\mathbf{X}}_i^{\top}\widetilde{\mathbf{X}}_i\right\|_F
\left\|\widetilde{\mathbf{X}}_{j}^{\top}\widetilde{\mathbf{X}}_{j}\right\|_F
},
\label{eq:expert_cka}
\end{equation}
where $\|\cdot\|_F$ denotes the Frobenius norm. This score is invariant to isotropic scaling and lies in $[0,1]$, with higher values indicating stronger alignment between the geometric patterns induced by the two experts.

\subsection{Rationale for Importance-Weighted Transformation}
\label{sec:importance_weighted_transformation_rationale}

Standard Tucker decomposition typically identifies low-rank structure by minimizing the Frobenius reconstruction error. Let $\mathbf W_i$ and $\widehat{\mathbf W}_i$ denote the original and compressed weights of expert $i$, respectively, and let $\boldsymbol\Delta_i=\mathbf W_i-\widehat{\mathbf W}_i$ denote the reconstruction error.
The standard objective is:
\begin{equation}
\min_{\widehat{\mathcal W}}
\left\|\mathcal W-\widehat{\mathcal W}\right\|_F^2
=
\min_{\widehat{\mathcal W}}
\sum_{i=1}^{E}\|\boldsymbol\Delta_i\|_F^2.
\label{eq:appendix_standard_tucker_error}
\end{equation}

Here, $\mathcal W$ and $\widehat{\mathcal W}$ are tensors obtained by stacking the original and compressed weights along the expert mode, and $E$ is the number of experts.
This objective penalizes reconstruction errors equally across weight directions.
However, input directions differ in activation energy, and output directions differ in their sensitivity to the task loss.
Consequently, weight errors of the same magnitude can have different effects on task performance.

The input second-moment matrix $\boldsymbol\Sigma_{\mathrm{act}}$ characterizes the energy of calibration inputs along different directions.
Let $\mathbf q_j^{\mathrm{act}}$ be a unit eigenvector of $\boldsymbol\Sigma_{\mathrm{act}}$, with corresponding eigenvalue $\lambda_j^{\mathrm{act}}$.
Then:
\begin{equation}
\begin{aligned}
\boldsymbol\Sigma_{\mathrm{act}}\mathbf q_j^{\mathrm{act}}
&=
\lambda_j^{\mathrm{act}}\mathbf q_j^{\mathrm{act}},
\mathbb E_{\mathrm{cal}}
\left[
\left((\mathbf q_j^{\mathrm{act}})^\top\mathbf A\right)^2
\right]
&=
\lambda_j^{\mathrm{act}}.
\end{aligned}
\label{eq:appendix_input_direction_energy}
\end{equation}

Here, $j$ indexes input directions, $\mathbf A$ is the projection input, and $\mathbb E_{\mathrm{cal}}$ denotes expectation over calibration samples.
A large $\lambda_j^{\mathrm{act}}$ indicates that calibration inputs have substantial energy along $\mathbf q_j^{\mathrm{act}}$, so weight errors along this direction can produce larger output deviations.
Conversely, when $\lambda_j^{\mathrm{act}}$ is close to zero, calibration inputs have small components along this direction, reducing its contribution to output reconstruction error.
Specifically:
\begin{equation}
\begin{aligned}
\mathbb E_{\mathrm{cal}}
\left[\|\boldsymbol\Delta_i\mathbf A\|_2^2\right]
&=
\operatorname{tr}
\left(
\boldsymbol\Delta_i^\top
\boldsymbol\Delta_i
\boldsymbol\Sigma_{\mathrm{act}}
\right)&=
\sum_{j=1}^{d_{\mathrm{in}}}
\lambda_j^{\mathrm{act}}
\left\|\boldsymbol\Delta_i\mathbf q_j^{\mathrm{act}}\right\|_2^2.
\end{aligned}
\label{eq:appendix_activation_weighted_error}
\end{equation}

Here, $d_{\mathrm{in}}$ is the input dimension and $\operatorname{tr}(\cdot)$ denotes the matrix trace.
This identity shows that the input second moment distinguishes reconstruction errors according to the energy of the corresponding input directions.

The importance of an output error further depends on its influence on the task loss.
The output gradient $\mathbf g_i=\partial\mathcal L_{\mathrm{cal}}/\partial\mathbf o_i$ and its second-moment matrix $\boldsymbol\Sigma_{\mathbf g}$, defined in the main text, characterize this sensitivity. Each output gradient $\mathbf g_i$ first forms its outer product $\mathbf g_i\mathbf g_i^\top$; these outer products are then accumulated over all routed token-expert records in the layer into the shared second-moment matrix $\boldsymbol\Sigma_{\mathbf g}$, which after normalization and shrinkage is Cholesky-factorized as $\widetilde{\boldsymbol\Sigma}_{\mathbf g}=\mathbf L_{\mathbf g}\mathbf L_{\mathbf g}^\top$, yielding the shared factor $\mathbf L_{\mathbf g}$.
Let $\mathbf q_{i,k}^{g}$ be a unit eigenvector of $\boldsymbol\Sigma_{\mathbf g}$, with corresponding eigenvalue $\lambda_{i,k}^{g}$.
Then:
\begin{equation}
\begin{aligned}
\boldsymbol\Sigma_{\mathbf g}\mathbf q_{i,k}^{g}
&=
\lambda_{i,k}^{g}\mathbf q_{i,k}^{g},
\mathbb E_{\mathrm{cal}}
\left[
\left((\mathbf q_{i,k}^{g})^\top\mathbf g\right)^2
\right]
&=
\lambda_{i,k}^{g}.
\end{aligned}
\label{eq:appendix_output_direction_sensitivity}
\end{equation}

Here, $k$ indexes output directions, $\mathcal L_{\mathrm{cal}}$ is the calibration loss, and $\mathbf o_i$ is the projection output.
Under a local first-order approximation of the loss, the directional derivative along $\mathbf q_{i,k}^{g}$ is $(\mathbf q_{i,k}^{g})^\top\mathbf g$.
Thus, a larger $\lambda_{i,k}^{g}$ indicates greater mean squared sensitivity along this output direction, motivating a stronger penalty on reconstruction errors in that direction.

Based on these observations, the transformation
$\widetilde{\mathbf W}_i
=
\mathbf L_{\mathbf g}^{\top}
\mathbf W_i
\mathbf L_{\mathrm{act}}$
incorporates both input energy and output sensitivity into the reconstruction metric.
Define
$\widehat{\widetilde{\mathbf W}}_i
=
\mathbf L_{\mathbf g}^{\top}
\widehat{\mathbf W}_i
\mathbf L_{\mathrm{act}}$.
Using the Cholesky factorizations
$\boldsymbol\Sigma_{\mathrm{act}}
=
\mathbf L_{\mathrm{act}}\mathbf L_{\mathrm{act}}^\top$
and
$\widetilde{\boldsymbol\Sigma}_{\mathbf g}
=
\mathbf L_{\mathbf g}\mathbf L_{\mathbf g}^\top$
from the main text, we obtain:
\begin{equation}
\begin{aligned}
\left\|
\widetilde{\mathbf W}_i-
\widehat{\widetilde{\mathbf W}}_i
\right\|_F^2
&=
\left\|
\mathbf L_{\mathbf g}^{\top}
\boldsymbol\Delta_i
\mathbf L_{\mathrm{act}}
\right\|_F^2
\\
&=
\operatorname{tr}\!\left(
\mathbf L_{\mathbf g}^{\top}
\boldsymbol\Delta_i
\mathbf L_{\mathrm{act}}
\mathbf L_{\mathrm{act}}^\top
\boldsymbol\Delta_i^\top
\mathbf L_{\mathbf g}
\right)
\\
&=
\operatorname{tr}\!\left(
\boldsymbol\Delta_i^\top
\widetilde{\boldsymbol\Sigma}_{\mathbf g}
\boldsymbol\Delta_i
\boldsymbol\Sigma_{\mathrm{act}}
\right).
\end{aligned}
\label{eq:appendix_weighted_metric_equivalence}
\end{equation}

Here, $\widetilde{\boldsymbol\Sigma}_{\mathbf g}$ is the normalized and shrunk gradient second-moment matrix used in the main text.
This transformation therefore changes the reconstruction metric in weight space: the ordinary Frobenius error in the transformed space is exactly equivalent to a weighted error in the original space that accounts for both input energy and task sensitivity.
We define this task-aware reconstruction error as:
\begin{equation}
\mathcal E_i
\left(\mathbf W_i,\widehat{\mathbf W}_i\right)
=
\operatorname{tr}\!\left[
(\mathbf W_i-\widehat{\mathbf W}_i)^\top
\widetilde{\boldsymbol\Sigma}_{\mathbf g}
(\mathbf W_i-\widehat{\mathbf W}_i)
\boldsymbol\Sigma_{\mathrm{act}}
\right].
\label{eq:appendix_task_aware_error}
\end{equation}

Here, $\mathcal E_i$ measures the weighted reconstruction error of expert $i$.
It places stronger penalties on errors along input directions with higher activation energy and output directions with greater loss sensitivity.
This quantity serves as a task-aware compression surrogate rather than an exact expression for the change in task loss.

Finally, second-moment weighting and Tucker factorization can be expressed as a unified optimization problem:
\begin{equation}
\begin{aligned}
\min_{\substack{
      \widetilde{\mathcal G},
      \widetilde{\mathbf U}_{\mathrm{exp}},\\
      \widetilde{\mathbf U}_{\mathrm{out}},
      \widetilde{\mathbf U}_{\mathrm{in}}}}
\quad&
\sum_{i=1}^{E}
\mathcal E_i
\left(\mathbf W_i,\widehat{\mathbf W}_i\right)
\\
\text{s.t.}\quad&
\widehat{\widetilde{\mathcal W}}
=
\widetilde{\mathcal G}
\times_1\widetilde{\mathbf U}_{\mathrm{exp}}
\times_2\widetilde{\mathbf U}_{\mathrm{out}}
\times_3\widetilde{\mathbf U}_{\mathrm{in}},
&
\widehat{\mathbf W}_i
=
\mathbf L_{\mathbf g}^{-\top}
\widehat{\widetilde{\mathbf W}}_i
\mathbf L_{\mathrm{act}}^{-1}.
\end{aligned}
\label{eq:appendix_task_aware_tucker_objective}
\end{equation}

Here, $\widehat{\widetilde{\mathcal W}}$ stacks the reconstructed transformed weights
$\widehat{\widetilde{\mathbf W}}_i$ along the expert mode.
The core tensor $\widetilde{\mathcal G}$ and the factor matrices
$\widetilde{\mathbf U}_{\mathrm{exp}}$,
$\widetilde{\mathbf U}_{\mathrm{out}}$, and
$\widetilde{\mathbf U}_{\mathrm{in}}$
follow the dimensions and compression ranks defined in the main text, and $\times_n$ denotes the mode-$n$ product.
The Tucker constraint is imposed in the weighted space.
Thus, second-moment weighting first determines which reconstruction errors matter more, and Tucker factorization then identifies shared low-rank structure across the expert, output-feature, and input-feature modes under this metric.

\subsection{Shared-factor Execution Strategy.}
\label{Computational_cost}

For a given MoE layer, let $\mathcal{E}$ be the set of $K$ activated experts, and let $\widehat{\pi}_i$ be the normalized routing weight of expert $\mathbf{E}_i$. The MoE output is:
\begin{equation}
\mathbf{A}_{\mathrm{out}}
=
\sum_{i\in\mathcal{E}}
\widehat{\pi}_i\,
\mathbf{W}_i^{\mathrm{down}}\,
\Big[
\phi\big(\mathbf{W}_i^{\mathrm{gate}}\mathbf{A}_{\mathrm{in}}\big)
\odot
\big(\mathbf{W}_i^{\mathrm{up}}\mathbf{A}_{\mathrm{in}}\big)
\Big],
\label{eq:cost_dense_forward}
\end{equation}
where $\mathbf{A}_{\mathrm{in}}$ is the input hidden state, $\phi$ is the element-wise activation, and $\odot$ denotes element-wise multiplication.
Using Eq.~\ref{eq:reconstruction}, the importance-weighted transformation can be absorbed into the factors as follows:
\begin{equation}
\hat{\mathbf U}_2=\mathbf{L}_{g}^{-\top}\widetilde{\mathbf{U}}_2,\qquad
\hat{\mathbf U}_1^\top=\widetilde{\mathbf{U}}_1^\top \mathbf{L}_{\mathrm{act}}^{-1},\qquad
\widetilde{\mathbf{G}}_i=\sum_{q=1}^{r_{exp}}[\widetilde{\mathbf{U}}_{\exp}]_{i,q}\widetilde{\mathbf{G}}^{(q)}.
\end{equation}

Therefore, given \(\mathbf{W}_i\in\mathbb{R}^{d_2\times d_1}\) corresponding to one of \(\mathbf{W}_i^{\mathrm{down}}\), \(\mathbf{W}_i^{\mathrm{gate}}\), or \(\mathbf{W}_i^{\mathrm{up}}\) for \(\mathrm{E}_i\), \(\mathbf{W}_i\) is decomposed as follows:
\begin{equation}
\mathbf{W}_i=\hat{\mathbf U}_2\widetilde{\mathbf{G}}_i \hat{\mathbf U}_1^\top,
\label{eq:cost_tucker_matrix}
\end{equation}
where \(\hat{\mathbf U}_1\in\mathbb{R}^{d_1\times r_1}\) is the input factor, \(\hat{\mathbf U}_2\in\mathbb{R}^{d_2\times r_2}\) is the output factor, and \(\widetilde{\mathbf{G}}_i\in\mathbb{R}^{r_2\times r_1}\) is the Tucker core.

\textbf{(1) Up projection and Gate projection.}
For up projection $\mathbf{W}^{\mathrm{up}}_i$, all activated experts receive the same $\mathbf{A}_{\mathrm{in}}$. Thus, by using Eq.~\ref{eq:cost_tucker_matrix}, the up projection is:
\begin{equation}
{\mathbf{W}}_i^{\mathrm{up}}
\mathbf{A}_{\mathrm{in}}
=
\hat{\mathbf U}_2^{\mathrm{up}}
\widetilde{\mathbf{G}}_i^{\mathrm{up}}
\left[
\left(\hat{\mathbf U}_1^{\mathrm{up}}\right)^{\top}
\mathbf{A}_{\mathrm{in}}
\right].
\label{eq:cost_up_forward}
\end{equation}
Since $\big(\hat{\mathbf U}_1^{\mathrm{up}}\big)^{\top}\mathbf{A}_{\mathrm{in}}$ is identical across experts, it is computed once and reused, reducing the number of input projections from $\mathcal{O}(K)$ to one.

The gate projection $\mathbf{W}^{\mathrm{gate}}_i$ also operates on the same $\mathbf{A}_{\mathrm{in}}$:
\begin{equation}
{\mathbf{W}}_i^{\mathrm{gate}}
\mathbf{A}_{\mathrm{in}}
=
\hat{\mathbf U}_2^{\mathrm{gate}}
\widetilde{\mathbf{G}}_i^{\mathrm{gate}}
\left[
\left(\hat{\mathbf U}_1^{\mathrm{gate}}\right)^{\top}
\mathbf{A}_{\mathrm{in}}
\right].
\label{eq:cost_gate_forward}
\end{equation}
Similarly, $(\hat{\mathbf U}_1^{\mathrm{gate}})^{\top}\mathbf{A}_{\mathrm{in}}$ is computed once instead of $K$ times. Thus, each expert then combines its gate and up outputs:
\begin{equation}
\mathbf{a}_i
=
\phi\!\left(
{\mathbf{W}}_i^{\mathrm{gate}}
\mathbf{A}_{\mathrm{in}}
\right)
\odot
\left(
{\mathbf{W}}_i^{\mathrm{up}}
\mathbf{A}_{\mathrm{in}}
\right).
\label{eq:cost_intermediate_activation}
\end{equation}

\textbf{(2) Down projection.} The down projection $\mathbf{W}^{\mathrm{down}}_i$ operates on the expert-specific activation $\mathbf{a}_i$:
\begin{equation}
\mathbf{W}_i^{\mathrm{down}}\mathbf{a}_i
=
\hat{\mathbf U}_2^{\mathrm{down}}\,
\widetilde{\mathbf{G}}_i^{\mathrm{down}}\,
\Big[\big(\hat{\mathbf U}_1^{\mathrm{down}}\big)^{\top}\mathbf{a}_i\Big],
\label{eq:cost_down_forward}
\end{equation}
where both $\hat{\mathbf U}_1^{\mathrm{down}}$ and $\hat{\mathbf U}_2^{\mathrm{down}}$ are shared. Although the input projection $\big(\hat{\mathbf U}_1^{\mathrm{down}}\big)^{\top}\mathbf{a}_i$ must be computed per expert, the shared output factor allows the final output to be evaluated as:
\begin{equation}
\mathbf{A}_{\mathrm{out}}
=
\hat{\mathbf U}_2^{\mathrm{down}}
\Big[
\sum_{i\in\mathcal{E}}
\widehat{\pi}_i\,
\widetilde{\mathbf{G}}_i^{\mathrm{down}}\,
\big(\hat{\mathbf U}_1^{\mathrm{down}}\big)^{\top}\mathbf{a}_i
\Big],
\label{eq:cost_down_shared_output}
\end{equation}
thereby reducing $K$ output projections to a single one.

\subsection{Rationale for Candidate-restricted Routing }
\label{why_cold}


In the SDAR-30B-A3B-Chat-b32 evaluation on GSM8K and MBPP calibration prompts, the average pairwise hidden-state cosine similarity of cold tokens is 0.4904, higher than the 0.2986 observed for hot tokens. This indicates that cold tokens are more directionally concentrated in hidden-state space, which is consistent with their more stable expert routing preferences. We formalize this property as follows.

Let $\mathbf h_i$ be the hidden state of the $i$-th token and define its
$L_2$-normalized representation as:
\begin{equation}
\mathbf x_i
=
\frac{\mathbf h_i}{\|\mathbf h_i\|_2},
\qquad
\|\mathbf x_i\|_2=1.
\label{eq:normalized-hidden}
\end{equation}

For a token set of size $N$, its average pairwise cosine similarity is:
\begin{equation}
S
=
\frac{1}{N(N-1)}
\sum_{i\neq j}
\mathbf x_i^\top \mathbf x_j.
\label{eq:cosine-similarity}
\end{equation}

Let
$\bar{\mathbf x}
=
\frac{1}{N}\sum_{i=1}^{N}\mathbf x_i$.
The average squared distance to the mean satisfies:
\begin{equation}
\frac{1}{N}
\sum_{i=1}^{N}
\left\|
\mathbf x_i-\bar{\mathbf x}
\right\|_2^2
=
\frac{N-1}{N}(1-S).
\label{eq:dispersion-similarity}
\end{equation}

Equivalently, defining the normalized directional dispersion as:
\begin{equation}
D
=
\frac{N}{N-1}
\frac{1}{N}
\sum_{i=1}^{N}
\left\|
\mathbf x_i-\bar{\mathbf x}
\right\|_2^2,
\end{equation}
we obtain$D=1-S.$ Therefore, the higher pairwise cosine similarity observed for cold tokens directly implies lower directional dispersion,
\begin{equation}
D_{\mathrm{cold}}
<
D_{\mathrm{hot}}.
\end{equation}

This directional concentration is consistent with the routing behavior
observed in our experiments: cold tokens exhibit more similar routing
probability distributions and larger Top-$K$ expert overlap than hot tokens.
Consequently, the centroid of cold-token hidden states provides a more
representative basis for constructing the candidate expert set, motivating
candidate-restricted routing for cold tokens.


\begin{table}[htbp]
\centering
\small
\renewcommand{\arraystretch}{1.1}

\setlength{\tabcolsep}{6pt}
\caption{Architectural configurations of the evaluated MoE-based diffusion language models.}
\label{tab:model_architectures}
\begin{tabular}{lcccc}
\toprule
Model 
& Layers 
& MoE Layers 
& Routed Experts 
& Experts/Token  \\
\midrule
SDAR-30B-A3B-Chat-b32
& 48 
& 48 
& 128 
& 8  \\
LLaDA2.0-mini
& 20 
& 19 
& 256 
& 8  \\
\bottomrule
\end{tabular}

\vspace{1.5em}

\setlength{\tabcolsep}{8pt}
\caption{Generation and diffusion decoding configurations used in our experiments.}
\label{tab:decoding_configurations}
\begin{tabular}{lcc}
\toprule
Configuration
& SDAR-30B-A3B-Chat-b32
& LLaDA2.0-mini \\
\midrule
Generation length
& 4096
& 4096 \\
Block length
& 32
& 32 \\
Denoising steps
& 32
& 32 \\
Confidence threshold
& 0.95
& 0.95 \\
Temperature
& 0.0
& 0.0 \\
Batch size
& 1
& 1 \\
\bottomrule
\end{tabular}

\vspace{1.5em}

\setlength{\tabcolsep}{6pt}
\caption{Hyperparameter settings of ITC-MoE.}
\label{ours}
\begin{tabular}{ccccc}
\toprule
$\lambda$ & $M$ & Compensation Rank & $\tau_h$ & $L_h$ \\
\midrule
0.1 & 48 & 11 & 0.7 & 3 \\
\bottomrule
\end{tabular}

\end{table}

\section{Backbone Model Details}
\label{app:backbone-Model-Details}

The architectural configurations of the evaluated MoE-based diffusion language models and the corresponding generation and diffusion decoding settings are summarized in Tables~\ref{tab:model_architectures} and~\ref{tab:decoding_configurations}, respectively. The hyperparameter settings of ITC-MoE are summarized in Table~\ref{ours}. Our method is implemented on top of the TEAM framework, upon which the proposed Importance-guided Adaptive Tucker Compression and Token-aware Compensation and Routing modules are further developed. For the Compensation Rank, a larger value generally yields better performance and we set it to 11 to balance performance and efficiency. $L_h$ follows the default setting in TEAM.

\section{Additional ablations}
\label{sec:Additional_ablations}

\begin{table}[t]
\centering
\small
\begin{minipage}[t]{0.48\textwidth}
\centering
\setlength{\tabcolsep}{6pt}
\captionof{table}{Effect of $\lambda$.}
\label{tab:lambda_ablation}
\begin{tabular}{ccc}
\toprule
$\lambda$ & MultiArith $\uparrow$ & SingleQ $\uparrow$ \\
\midrule
0.0 & 98.33 & 93.58 \\
0.1 & 99.17 & 97.25 \\
0.2 & 98.83 & 95.41 \\
0.3 & 98.67 & 94.50 \\
\bottomrule
\end{tabular}
\end{minipage}
\hfill
\begin{minipage}[t]{0.48\textwidth}
\centering
\setlength{\tabcolsep}{6pt}
\captionof{table}{Sensitivity to $M$.}
\label{tab:m_ablation}
\begin{tabular}{ccc}
\toprule
$M$ & SingleOp (\%) & GSM8K (\%) \\
\midrule
16 & 93.24 & 85.52 \\
32 & 94.48 & 86.96 \\
48 & 95.37 & 87.87 \\
64 & 95.37 & 88.02 \\
\bottomrule
\end{tabular}
\end{minipage}
\end{table}
 
\textbf{Effect of the Compensation Regularization Coefficient.}
We investigate the sensitivity of the token-aware compensation module to the regularization coefficient $\lambda$ in Eq.~\ref{eq:hot_compensation_loss}. As shown in Table~\ref{tab:lambda_ablation}, under a 10\% compression ratio on SDAR, $\lambda=0.1$ consistently achieves the best performance on both MultiArith and SingleQ. A smaller $\lambda$ provides insufficient regularization and may lead to over-compensation, whereas a larger $\lambda$ overly constrains the compensation module and limits its ability to correct compression-induced errors. Therefore, we set $\lambda=0.1$ as the default value in all experiments.

\textbf{Effect of Candidate Set Size \(M\).}
We further investigate the effect of the candidate set size \(M\) on model performance. As shown in Table~\ref{tab:m_ablation}, the performance on SingleOp and GSM8K generally improves as \(M\) increases. When \(M\) reaches 48, the performance gains begin to saturate. Therefore, we set \(M=48\) as the default configuration to balance model performance and routing efficiency.

\end{document}

%% file: math_commands.tex
\usepackage{amsmath,amsfonts,bm}

\def\eqref#1{equation~\ref{#1}}

\def\1{\bm{1}}

\DeclareMathAlphabet{\mathsfit}{\encodingdefault}{\sfdefault}{m}{sl}
\SetMathAlphabet{\mathsfit}{bold}{\encodingdefault}{\sfdefault}{bx}{n}



%% file: paper/introduct.tex
\section{Introduction}
\label{sec:introduction}

Mixture-of-Experts (MoE)-based Diffusion Language Models (DLMs) have recently emerged as a promising alternative to autoregressive language models~\citep{Diffusion-LM,MaskDLLM}. By generating sequences through multiple denoising iterations, MoE DLMs increase model capacity with multiple experts and selectively activate a subset of experts for each token~\citep{zhu2026lladamoev2scaling}. Unlike autoregressive models, DLMs can predict tokens at multiple positions in parallel within each iteration, enabling flexible parallel decoding and strong performance across various tasks~\citep{LLDM,Faster-Than-AR}.

However, the large number of expert parameters in MoE models introduces substantial computation and storage overhead, becoming a critical bottleneck for practical DLM inference~\citep{team2026}.
Low-rank compression provides a natural approach to reducing the computation and storage costs of MoE models. Existing methods~\citep{li2025moesvd,gu2025d2moe,chen2026mobe,mi2026rfid} have demonstrated the effectiveness of low-rank factorization for conventional autoregressive MoE models. However, their compression strategies are not well suited to the distinctive characteristics of MoE DLMs. As shown in Fig.~\ref{fig:globalmotivation}, we identify two properties of MoE DLMs: \emph{cross-mode non-uniform redundancy} and \emph{token-wise utilization variation}.

\begin{figure}
    \centering
    \includegraphics[width=0.9\linewidth]{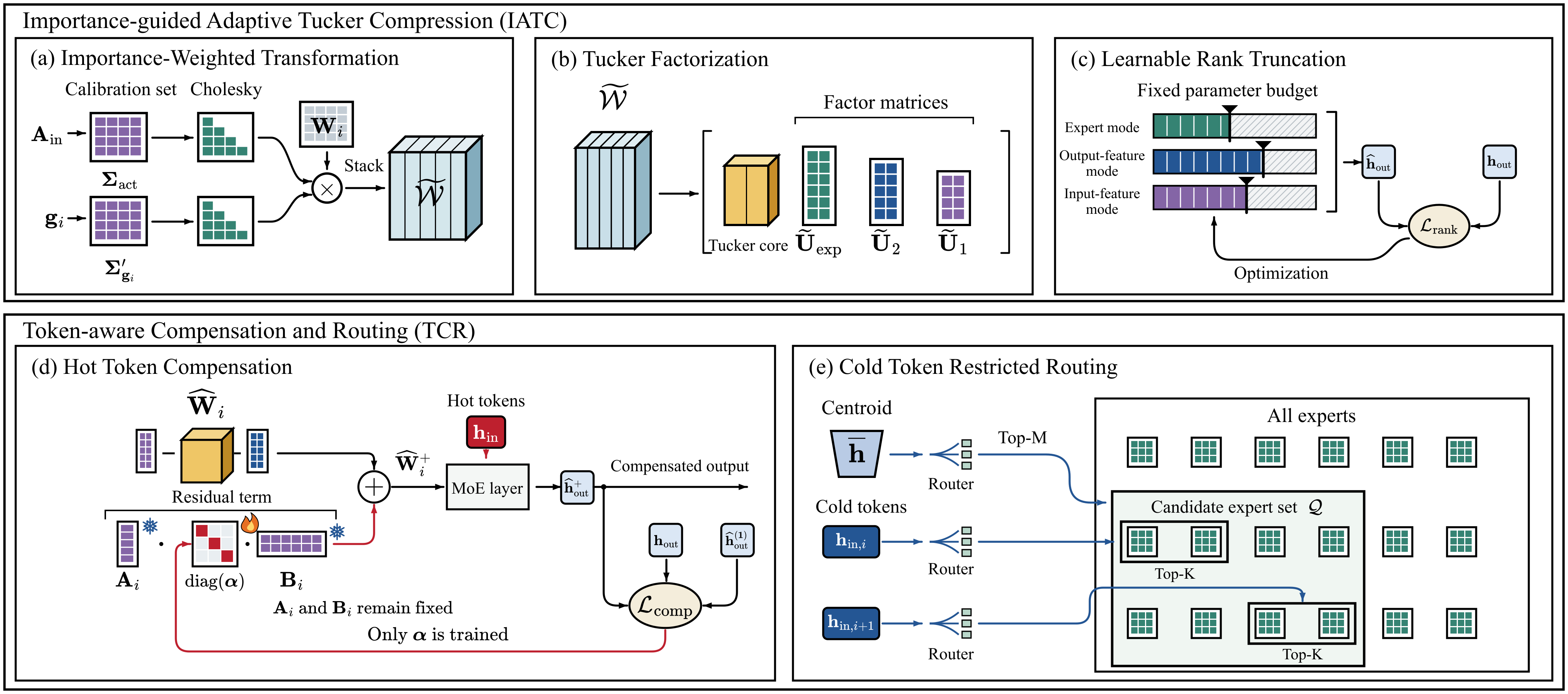}
    \caption{Overview of the ITC-MoE framework.}
    \label{fig:famework}
    \vspace{-5mm}
\end{figure}

First, as shown in Fig.~\ref{fig:globalmotivation}(a)-(e), the redundancy of MoE parameters is inherently non-uniform across modes. The input and output feature dimensions exhibit strong low-rank structures, while the expert dimension also contains substantial cross-expert redundancy, and different modes show distinct sensitivities to rank truncation. 
Second, during diffusion decoding, MoE DLMs exhibit pronounced token-wise utilization variation. Hot and cold tokens differ in their feature spectral distributions and expert utilization patterns, with hot tokens being more sensitive to rank truncation and cold tokens exhibiting more concentrated expert activation. 
However, existing low-rank MoE compression methods~\citep{li2025moesvd,gu2025d2moe,chen2026mobe,mi2026rfid,team2026} largely rely on static compression schemes, such as independent expert factorization or pre-determined rank and routing configurations. Such approaches neither explicitly exploit the non-uniform redundancy across the input, output, and expert dimensions nor adapt the compressed model to token-wise variation during diffusion decoding~\citep{huang2026dont,jazbec2026learning}, leaving opportunities for further reducing computation and storage of MoE DLMs while preserving model capacity.

In this paper, we propose \textbf{ITC-MoE}, an Importance-guided Token-aware Compression framework for MoE DLMs, as shown in Fig.~\ref{fig:famework}. ITC-MoE jointly exploits non-uniform parameter redundancy and token-wise utilization variation. 
At first, ITC-MoE introduces Importance-guided Adaptive Tucker Compression (IATC) to exploit cross-mode non-uniform redundancy. IATC incorporates activation and gradient importance into expert weight transformation, jointly factorizes the transformed expert weights with Tucker decomposition, and adaptively allocates ranks across different modes under a fixed parameter budget. 
ITC-MoE then introduces Token-aware Compensation and Routing (TCR) to adapt the compressed experts to token-wise variation. TCR applies lightweight low-rank compensation to hot tokens, which are more sensitive to compression errors, while restricting the candidate expert set for cold tokens, whose routing decisions are more concentrated. 

By jointly integrating IATC and TCR, ITC-MoE enables more efficient MoE DLM inference while maintaining model fidelity. For example, on SDAR-30B-A3B-Chat-b32~\cite{sdar}, under a 30\% compression budget, ITC-MoE still achieves an accuracy of 96.33 on MultiArith while delivering up to a 7.22$\times$ end-to-end inference speedup.

%% file: paper/related_work.tex
\section{Related Work}

\textbf{Mixture-of-Experts for Diffusion Models.}
The iterative nature of diffusion models naturally aligns with MoE routing, as the computational demands vary across timesteps and token positions, enabling dynamic expert assignment.
Switch-DiT~\citep{park2024switchdit} first introduces sparse MoE into Diffusion Transformers, learning timestep-dependent expert paths. EC-DiT~\citep{sun2025ecdit} adopts expert-choice routing to adaptively allocate computation according to heterogeneous text and image-patch complexity. Diff-MoE~\citep{cheng2025diffmoe} introduces time-aware and space-adaptive experts, conditioning expert assignment jointly on timestep and spatial token features. Race-DiT~\citep{yuan2025racedif} formulates routing as a competition between tokens and experts, dynamically assigning additional expert capacity to important tokens.
Complementary to these routing-centric approaches, Dense2MoE~\citep{zheng2025dense2moe} converts pretrained dense diffusion transformers into sparse MoE architectures with block-level conditional execution. 
ProMoE~\citep{wei2026promoe} argues that conventional language-model routing does not directly transfer to visual diffusion models, and proposes explicitly guided conditional and prototype-based routing to promote expert specialization. Beyond architectural innovations, MoE has also been incorporated into compute-efficient training pipelines to scale model capacity under constrained budgets~\citep{sehwag2025microdiffusion}.

\textbf{Efficient Mixture-of-Experts.}
MoE models dynamically route each token to a small subset of experts, which incurs considerable computational overhead. Many efficient research methods have been developed to improve routing strategies and expert utilization~\cite{dai2024deepseekmoe,lewis2021base,zhou2022expertchoice,komatsuzaki2023sparse}. 
%
Soft MoE~\citep{puigcerver2024softmoe} replaces discrete routing with differentiable soft assignment, improving training stability and expert utilization. 
%
Another line of work reduces the memory and computational cost of experts by employing quantization~\cite{xu2026kbvq,chen2025moequant,duanmu2025mxmoe}. 
%
%
Expert-level pruning methods identify and remove redundant experts~\citep{jie2025mole,huang2025shapleymoe,bai2025diep}. 
FloE~\citep{zhou2025floe} combines compression with on-demand parameter loading for memory-constrained inference, and Oracle-MoE~\citep{zhou2025oraclemoe} improves routing locality to reduce expert swapping. 
Recent methods explicitly target low-rank redundancy within experts.
MoBE~\citep{chen2026mobe} represents experts using shared basis matrices. MoNE~\citep{zhang2026mone} replaces redundant experts with lightweight novice modules. 
TD-MoE~\citep{xu2026tdmoe} performs Tucker decomposition to capture cross-expert structure.
MiLo~\citep{huang2025milo} combines aggressive expert quantization with low-rank compensators.
MoE-SVD~\citep{li2025moesvd} shares decomposed components across experts to reduce both parameters and latency.
%
%
However, these low-rank methods target autoregressive MoE LLMs and have not been adapted for MoE DLMs, which use bidirectional attention.
As a recent effort, TEAM~\citep{team2026} addresses this gap by exploiting spatiotemporal consistency in MoE DLM decoding to adaptively activate experts according to token states, thereby reducing redundant computation. However, TEAM only reduces inference overhead, without fully exploiting the structural redundancy in experts to compress the model footprint.

%% file: paper/method.tex
\section{Method}
\label{sec:method}
 
\subsection{Preliminaries}
\label{sec:prelim}

\textbf{Mixture-of-Experts.}
A mixture-of-experts (MoE) layer replaces a dense feed-forward network with $E$ experts and a router~\citep{shazeer2017}. For a input $\mathbf{h}_{\mathrm{in}}\in\mathbb{R}^{d_{\mathrm{in}}}$, the router computes
$
\boldsymbol{\pi}(\mathbf{h}_{\mathrm{in}})
=
\operatorname{softmax}\!\left(
\mathbf{W}^{\boldsymbol{\pi}}\mathbf{h}_{\mathrm{in}}
\right)
\label{eq:moe_router}
$, where $\mathbf{W}^{\boldsymbol{\pi}}\in\mathbb{R}^{E\times d_{\mathrm{in}}}$ is the router weight and $\pi_i(\mathbf{h}_{\mathrm{in}})$ is the routing probability for expert $\mathbf{E}_i$. The selected expert set is $\mathcal{E}=\operatorname{Top}_{K}(\boldsymbol{\pi}(\mathbf{h}_{\mathrm{in}}))$, where $\operatorname{Top}_{K}(\cdot)$ returns the indices of the $K$ highest-scoring experts. For a gated feed-forward expert, the output of expert $\mathbf{E}_i$ for the input $\mathbf{h}_\mathrm{in}$ is:
\begin{equation}
\mathbf{o}_{i}
=
\mathbf{W}^{\mathrm{down}}_i
\left[
\phi(\mathbf{W}^{\mathrm{gate}}_i\mathbf{h}_{\mathrm{in}})
\odot
(\mathbf{W}^{\mathrm{up}}_i\mathbf{h}_{\mathrm{in}})
\right],
\label{eq:moe_expert_output}
\end{equation}
where $\mathbf{W}^{\mathrm{down}}_i,\mathbf{W}^{\mathrm{gate}}_i,\mathbf{W}^{\mathrm{up}}_i$ are the weights of $\mathbf{E}_i$, $\phi(\cdot)$ is an element-wise activation, and $\odot$ denotes element-wise multiplication. The MoE output is:
\begin{equation}
\mathbf{h}_{\mathrm{out}}
=
\sum_{i\in\mathcal{E}}
\widehat{\pi}_{i}\,
\mathbf{o}_{i},
\qquad
\widehat{\pi}_{i}
=
\frac{
\pi_i(\mathbf{h}_{\mathrm{in}})
}{
\sum_{j\in\mathcal{E}}
\pi_j(\mathbf{h}_{\mathrm{in}})
},
\label{eq:moe_output}
\end{equation}
where $\widehat{\pi}_{i}$ is the routing weight of the $i$th expert after renormalization over $\mathcal{E}$ and $\mathbf{h}_{\mathrm{out}}\in\mathbb{R}^{d_{\mathrm{out}}}$ is the aggregated output.

\textbf{Hot and Cold Tokens in DLMs.}
During diffusion decoding, currently masked positions are classified as hot state or cold state according to their likelihood of being accepted in subsequent decoding steps~\citep{team2026}.
At diffusion step $t$, let $\mathcal{M}_t$ and $\mathcal{D}_t$ denote the sets of masked and accepted token positions, respectively. The hot token set is:
\begin{equation}
\mathcal{H}_t
=
\left\{
i\in\mathcal{M}_t
\;\middle|\;
c_{i,t}>\tau_h
\ \text{or}\
\exists q\in\mathcal{D}_t:|i-q|\leq L_h
\right\},
\label{eq:hot}
\end{equation}
where $i$ and $q$ index masked and accepted token positions, respectively, $c_{i,t}$ is the confidence at position $i$, $\tau_h$ is the confidence threshold, and $L_h$ is the neighborhood radius.
The confidence is:
\begin{equation}
c_{i,t}
=
\max_{v\in\mathcal{V}}
\Pr_{\theta}\!\left(y_i=v\mid\mathbf{x},\mathbf{Y}_t\right).
\label{eq:confidence}
\end{equation}
The vocabulary is denoted by $\mathcal{V}$, the candidate token by $v$, the model parameters by $\theta$, and the token at position $i$ by $y_i$. The conditioning input is denoted by $\mathbf{x}$, and the partially masked target sequence at step $t$ is denoted by $\mathbf{Y}_t$.
In particular, $\mathbf{x}$ is the conditioning input, and $\mathbf{Y}_t$ is the partially masked sequence.
The remaining masked positions form the cold token set $\mathcal{C}_t=\mathcal{M}_t\setminus\mathcal{H}_t$.

\begin{figure*}[ht]
    \centering
    \includegraphics[width=1\linewidth]{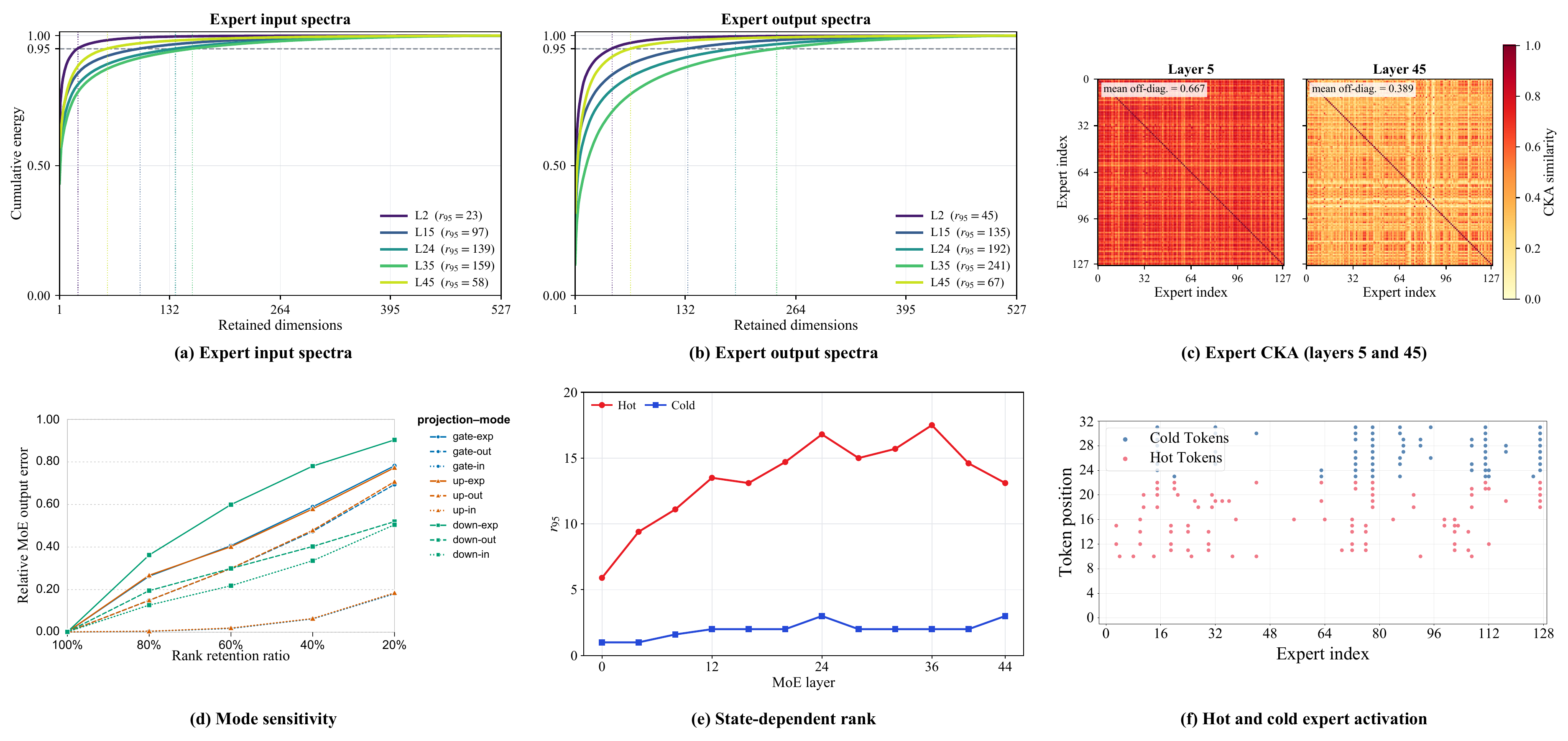}
    \vspace{-2mm}
    \caption{
\textbf{Observations.}
Results on SDAR-30B-A3B-Chat-b32~\citep{sdar}. Panels (a)-(d) use the MBPP benchmark, while (e) and (f) use the GSM8K benchmark. \textbf{(a,b) Expert weight spectra.} Cumulative spectral energy of the layerwise expert-weight tensor along input- and output-feature modes, respectively. \textbf{(c) Expert representation similarity.} Pairwise linear CKA between expert output representations within a layer, where higher values indicate higher similarity. ``mean off-diag.'' is the mean off-diagonal CKA. \textbf{(d) Mode sensitivity.} Relative MoE output reconstruction error \emph{vs.} dense reference when each Tucker mode is truncated independently with the other two at full rank. \textbf{(e) Token-dependent rank.} Ranks retaining 95\% energy for hot and cold tokens across MoE layers. \textbf{(f) Token-dependent expert patterns.} Distribution of experts activated by cold and hot tokens under Top-8 routing.
}
\vspace{-0.7cm}
\label{fig:globalmotivation}
\end{figure*}

\subsection{Motivation}
\label{motivation}

As shown in Fig.~\ref{fig:globalmotivation}, we identify two key properties of MoE DLMs: \emph{cross-mode non-uniform redundancy} and \emph{token-wise utilization variation}.
Specifically, Fig.~\ref{fig:globalmotivation}(a) and (b) show that the spectra of both input and output modes concentrate most of their energy in a few dominant directions, indicating a strong low-rank structure. Fig.~\ref{fig:globalmotivation}(c) further presents the linear CKA values~\citep{CKA} between paired experts, which quantify the similarity of their output feature geometries and thus reflect structural redundancy across experts (see Appendix~\ref{sec:app-cka} for details). The observed repeated patterns across experts confirm the presence of cross-expert redundancy. Fig.~\ref{fig:globalmotivation}(d) further shows that different modes exhibit varying sensitivity to rank truncation. 
Together, these results reveal substantial cross-mode non-uniform redundancy in MoE DLMs, suggesting that effective compression should adaptively exploit redundancy across all modes.
Moreover, as shown in Fig.~\ref{fig:globalmotivation}(e), hot-token features exhibit broader energy distributions, whereas cold tokens concentrate their energy in fewer ranks, indicating that hot tokens are more sensitive to rank truncation. Fig.~\ref{fig:globalmotivation}(f) also shows that, compared with hot tokens, cold tokens generally concentrate their activated experts on a smaller set.
These results indicate substantial token-wise utilization variation within MoE DLMs, suggesting that effective compression should be tailored to different token states.
 
However, existing low-rank MoE compression methods~\citep{li2025moesvd,gu2025d2moe,chen2026mobe,mi2026rfid,team2026} fail to jointly accommodate these two properties. They either factorize each expert independently without explicitly exploiting non-uniform redundancy across all of the input, output, and expert modes, or pre-determine rank budgets and expert routing statically and uniformly. Such static schemes cannot adapt to the token-wise utilization variation that emerges dynamically during the denoising process of MoE DLMs~\citep{huang2026dont,jazbec2026learning}, limiting their effectiveness and efficiency.

\subsection{Importance-guided Token-aware Compression for MoE DLMs} 
  
As shown in Fig.~\ref{fig:famework}, the proposed Importance-guided Token-aware Compression (ITC-MoE) contains two components, including importance-guided Adaptive Tucker Compression (IATC) and token-aware Compensation and Routing (TCR).

\subsubsection{Importance-guided Adaptive Tucker Compression}

To adapt cross-mode non-uniform redundancy, as shown in Fig.~\ref{fig:famework}, ITC-MoE introduces Importance-guided Adaptive Tucker Compression (IATC), which consists of three sequential steps, including Importance-Weighted Transformation, Tucker Factorization, and Learnable Rank Truncation.

\textbf{Importance-Weighted Transformation.}
For the $i$-th expert $\mathbf{E}_i$, we consider a weight matrix $\mathbf{W}_i \in \mathbb{R}^{d_2 \times d_1}$ that may be $\mathbf{W}^{\mathrm{down}}_i$, $\mathbf{W}^{\mathrm{gate}}_i$, or $\mathbf{W}^{\mathrm{up}}_i$. Its input and output activations are denoted by $\mathbf{A}_{\mathrm{in}} \in \mathbb{R}^{d_1}$ and $\mathbf{A}_{\mathrm{out}} \in \mathbb{R}^{d_2}$, respectively, and the model loss over the calibration set is denoted by $\mathcal{L}_{\mathrm{cal}}$. We define the output gradient, activation second-moment matrix, and gradient second-moment matrix as:
\begin{equation}
\mathbf{g}
=
 \partial\mathcal{L}_{\mathrm{cal}}/\partial\mathbf{A}_{\mathrm{out}},
\qquad
\boldsymbol{\Sigma}_{\mathrm{act}}
=
\mathbb{E}_{\mathrm{cal}}
\!\left[
\mathbf{A}_{\mathrm{in}}\mathbf{A}_{\mathrm{in}}^{\top}
\right],
\qquad
\boldsymbol{\Sigma}_{\mathbf{g}}
=
\mathbb{E}_{\mathrm{cal}}
\!\left[
\mathbf{g}\mathbf{g}^{\top}
\right],
\label{eq:symbolic_covariance}
\end{equation}
where $\mathbb{E}_{\mathrm{cal}}[\cdot]$ aggregates statistics over calibration tokens for the corresponding projection. Here, $\boldsymbol{\Sigma}_{\mathrm{act}}$ captures frequently activated input directions, while $\boldsymbol{\Sigma}_{\mathbf{g}}$ captures loss-sensitive output directions. To avoid ill-conditioning of the gradient statistics, we use the shrinkage estimator~\citep{ollila2024linear}:
\begin{equation}
\boldsymbol{\Sigma}'_{\mathbf{g}}
=
(1-\eta)\mathbf{I}_{d_2}
+
\eta
\frac{
\boldsymbol{\Sigma}_{\mathbf{g}}
}{
\operatorname{tr}(\boldsymbol{\Sigma}_{\mathbf{g}})/d_2
},
\label{eq:symbolic_shrinkage}
\end{equation}
where $\mathbf{I}_{d_2}$ is the $d_2 \times d_2$ identity matrix and $\eta\in[0,1]$ controls the shrinkage strength. Let $\mathbf{L}_{\mathrm{act}}$ and $\mathbf{L}_{\mathbf{g}}$ be the Cholesky inverses factors~\citep{lu2024numericalmatrixdecomposition} satisfying $\boldsymbol{\Sigma}_{\mathrm{act}}=\mathbf{L}_{\mathrm{act}}\mathbf{L}_{\mathrm{act}}^{\top}$ and $\boldsymbol{\Sigma}'_{\mathbf{g}}=\mathbf{L}_{\mathbf{g}}\mathbf{L}_{\mathbf{g}}^{\top}$. We then transform $\mathbf{W}_i$ into an importance-weighted representation $\widetilde{\mathbf{W}}_i
=
\mathbf{L}_{\mathbf{g}}^{\top}
\mathbf{W}_i
\mathbf{L}_{\mathrm{act}}$. 
This emphasizes reconstruction along frequently activated input and loss-sensitive output directions, yielding a suitable representation for low-rank factorization~\citep{witer,Self-Supervised_learn}. See Appendix~\ref{sec:importance_weighted_transformation_rationale} for theoretical analysis.

\textbf{Tucker Factorization.}
Then, for each MoE layer, we stack the importance-weighted expert matrices along the expert mode to form $\widetilde{\mathcal{W}}=\operatorname{stack}\nolimits_{i=1}^{E}(\widetilde{\mathbf{W}}_i)\in\mathbb{R}^{E\times d_2\times d_1}$, where $E$ is the number of experts. The resulting tensor is factorized by Tucker factorization~\citep{take_Y,takeA}:
\begin{equation}
\widetilde{\mathcal{W}}
\approx
\widetilde{\mathcal{G}}
\times_1
\widetilde{\mathbf{U}}_{\mathrm{exp}}
\times_2
\widetilde{\mathbf{U}}_2
\times_3
\widetilde{\mathbf{U}}_1,
\label{eq:symbolic_tucker}
\end{equation}
where $\times_n$ denotes the mode-$n$ tensor-matrix product and the Tucker core $\widetilde{\mathcal{G}}\in \mathbb{R}^{r_{\mathrm{exp}}\times r_2\times r_1}$ captures interactions among modes. The factor matrices
$
\widetilde{\mathbf{U}}_{\mathrm{exp}}
\in\mathbb{R}^{E\times r_{\mathrm{exp}}},
\widetilde{\mathbf{U}}_2
\in\mathbb{R}^{d_2\times r_2},
\widetilde{\mathbf{U}}_1
\in\mathbb{R}^{d_1\times r_1}
$
correspond to the expert, output-feature, and input-feature modes, with retained ranks $r_{\mathrm{exp}}$, $r_2$, and $r_1$. Setting $r_{\mathrm{exp}}<E$, $r_2<d_2$, and $r_1<d_1$ reduces both the parameter footprint and computational overhead. The reconstructed weight in the original space is:
\begin{equation}
\widehat{\mathbf{W}}_i
=
\mathbf{L}_{\mathbf{g}}^{-\top}
\widetilde{\mathbf{U}}_2
\left(
\sum\nolimits_{q=1}^{r_{\mathrm{exp}}}
[\widetilde{\mathbf{U}}_{\mathrm{exp}}]_{i,q}
\widetilde{\mathcal{G}}^{(q)}
\right)
\widetilde{\mathbf{U}}_1^{\top}
\mathbf{L}_{\mathrm{act}}^{-1},
\label{eq:reconstruction}
\end{equation}
where $[\widetilde{\mathbf{U}}_{\mathrm{exp}}]_{i,q}$ is the coefficient of the $q$-th expert-mode component for expert $\mathbf{E}_i$, and $\widetilde{\mathcal{G}}^{(q)}$ is the $q$-th matrix slice of $\widetilde{\mathcal{G}}$ along the expert mode. Since $\widetilde{\mathbf{U}}_1$, $\widetilde{\mathbf{U}}_2$, and the Cholesky inverses are fixed during inference, they can be fused offline and stored in advance, avoiding additional online computation.

\textbf{Learnable Rank Truncation.}
As shown in Fig.~\ref{fig:globalmotivation}(d), different modes exhibit varying sensitivity to rank truncation, making uniform rank allocation suboptimal for balancing compression efficiency and fidelity. Motivated by this, we therefore propose Learnable Rank Truncation, which adaptively allocates the rank budget across modes by minimizing output reconstruction error.
Consider a weight tensor $\widetilde{\mathcal{W}}\in\mathbb{R}^{E\times d_2\times d_1}$ with Tucker ranks $r_a$, $a\in\{\mathrm{exp},2,1\}$. Under a fixed parameter budget $N_{\mathrm{param}}
=
r_{\mathrm{exp}}r_2r_1
+
Er_{\mathrm{exp}}
+
d_2r_2
+
d_1r_1$, the discrete ranks cannot be directly optimized due to non-differentiable truncation. We therefore relax each rank into a continuous truncation position:
\begin{equation}
\rho_a
=
r_a^{\mathrm{lower}}
+
\left(
r_a^{\mathrm{upper}}
-
r_a^{\mathrm{lower}}
\right)
\sigma\!\left(\vartheta_a+\xi\right),
\label{eq:normalized_rank_allocation}
\end{equation}
where $r_a^{\mathrm{lower}}$ and $r_a^{\mathrm{upper}}$ are the minimum and maximum ranks of mode $a$, respectively. The learnable parameter $\vartheta_a$ determines the relative rank allocation across modes, while the shared offset $\xi$ is determined via binary search to satisfy the fixed budget $N_{\mathrm{param}}$. A differentiable soft mask for the $k$-th component of mode $a$ is given by $
m_{a,k}
=
\sigma\!\left(
\frac{\rho_a-k}{\tau}
\right),
\,
k=1,\ldots,D_a$, where $D_a$ is the dimension size of mode $a$ and $\tau$ is the temperature. The mask preserves components before $\rho_a$ and suppresses those beyond it, providing a differentiable approximation to hard truncation. The temperature $\tau$ is gradually annealed during optimization, causing the soft masks to approach hard truncation.
The rank allocation is optimized with the reconstruction loss:
\begin{equation}
\mathcal{L}_{\mathrm{rank}}
=
(
\mathbf{h}_{\mathrm{out}}
-
\widehat{\mathbf{h}}_{\mathrm{out}}
)^{\top}
\mathbf{H}
(
\mathbf{h}_{\mathrm{out}}
-
\widehat{\mathbf{h}}_{\mathrm{out}}
),
\label{eq:rank_allocation_loss}
\end{equation}
where $\mathbf{h}_{\mathrm{out}}$ and $\widehat{\mathbf{h}}_{\mathrm{out}}$ are the dense and softly truncated outputs, respectively, and $\mathbf{H}$ is the trace-normalized second-moment matrix of output gradients aggregated over all experts. Minimizing $\mathcal{L}_{\mathrm{rank}}$ thus allocates more rank capacity to modes that are more sensitive to truncation.

\textbf{Shared-factor Execution Strategy.} 
This factorization enables a shared-factor execution strategy that accelerates runtime. For up and gate projections, the input factors are shared across experts since all activated experts receive the same input. Thus, the shared input projections are computed once and reused. For the down projection, both input and output factors are shared. Although the intermediate activations differ, the per-expert weighted contributions can be aggregated in the low-rank core space, followed by a single shared output projection. This lowers the computational cost of both input and output projections while preserving the compressed MoE formulation. In our experiments, we found that this strategy achieves up to a $2.18\times$ speedup. Full details are in Appendix~\ref{Computational_cost}.

\subsubsection{Token-aware Compensation and Routing}

Then, to adapt the token-wise utilization variation, as shown in Fig.~\ref{fig:famework}, ITC-MOE introduces Token-aware Compensation and Routing (TCR), including Hot Token Compensation and Cold Token Restricted Routing.

\textbf{Hot Token Compensation.}
To protect hot tokens likely retained in later decoding, we introduce trainable compensation. For hot tokens, we augment each truncated Tucker reconstructed weight $\widehat{\mathbf{W}}_i$ with a residual term $\widehat{\mathbf{W}}_i^{+}
=
\widehat{\mathbf{W}}_i
+
\mathbf{A}_i\operatorname{diag}\!\left(\boldsymbol{\alpha}\right)\mathbf{B}_i$,
where $\mathbf{A}_i\in\mathbb{R}^{d_2\times r_{\mathrm{res}}}$ and $\mathbf{B}_i\in\mathbb{R}^{r_{\mathrm{res}}\times d_1}$ span the compensation subspace, $r_{\mathrm{res}}$ is the residual rank, and $\boldsymbol{\alpha}\in\mathbb{R}^{r_{\mathrm{res}}}$ are learnable parameters. As the initialization, we set $\boldsymbol{\alpha}=\mathbf{1}$ and solve $\boldsymbol{\alpha}, \mathbf{A}_i,\mathbf{B}_i$ by:
\begin{equation}
(\mathbf{A}_i^{*},\mathbf{B}_i^{*})
=
\arg\min_{\mathbf{A}_i,\mathbf{B}_i}
\left\|
\left(
\mathbf{W}_i
-
\widehat{\mathbf{W}}_i
-
\mathbf{A}_i\mathbf{B}_i
\right)
\mathbf{L}_{\mathrm{act}}
\right\|_F^2,
\label{eq:hot_weighted_residual}
\end{equation}
where $\mathbf{L}_{\mathrm{act}}$ is the Cholesky factor of the input second-moment matrix estimated from hot tokens, emphasizing input directions frequently activated by hot tokens. After initialization, $\mathbf{A}_i$ and $\mathbf{B}_i$ remain fixed, and only $\boldsymbol{\alpha}$ is trained with the following objective:
\begin{equation}
\begin{aligned}
\mathcal{L}_{\mathrm{comp}}
={}&
\left\|
\widehat{\mathbf{h}}_{\mathrm{out}}^{+}
-
\mathbf{h}_{\mathrm{out}}
\right\|_2^2
+
\lambda
\left\|
\widehat{\mathbf{h}}_{\mathrm{out}}^{+}
-
\widehat{\mathbf{h}}_{\mathrm{out}}^{(\mathbf{1})}
\right\|_2^2,
\end{aligned}
\label{eq:hot_compensation_loss}
\end{equation}
where $\mathbf{h}_{\mathrm{out}}$ is the original uncompressed MoE-layer output, $\widehat{\mathbf{h}}_{\mathrm{out}}^{+}$ is the compensated output, 
\begin{wraptable}{r}{0.45\textwidth}
\centering
\vspace{-2mm}
\caption{Comparison of Top-8 recall for different candidate set sizes on cold and hot tokens. Top-8 recall is defined as the fraction of tokens whose ground-truth Top-8 expert set is fully contained in the candidate set. Evaluation uses the SDAR-30B-A3B-Chat-b32 model with held-out GSM8K and MBPP calibration prompts, with 128 experts and Top-8 routing per token.
}
\vspace{-2mm}
\label{tab:candidate-recall}
\begin{tabular}{lcc}
\toprule
$M$ & Cold recall & Hot recall \\
\midrule
16 & 0.8975 & 0.6930 \\
32 & 0.9627 & 0.8186 \\
48 & 0.9795 & 0.8834 \\
64 & 0.9890 & 0.9257 \\
\bottomrule
\end{tabular}
\vspace{-6mm}
\end{wraptable}
and $\widehat{\mathbf{h}}_{\mathrm{out}}^{(\mathbf{1})}$ is the fixed reference output at $\boldsymbol{\alpha}=\mathbf{1}$. The first term minimizes reconstruction error relative to the original MoE-layer output, recovering truncation-induced information loss. The second anchors the compensated output to its initialization, preventing overfitting and stabilizing optimization. The coefficient $\lambda$ controls regularization strength.

\textbf{Cold Token Restricted Routing.}
During diffusion decoding, numerous masked tokens repeatedly undergo expert routing, activating many distinct experts within a single forward pass and increasing expert weight memory access and execution overhead~\citep{team2026,Klotski}.
Fortunately, cold tokens generally concentrate their activated experts on a small set, indicating that full-expert routing is unnecessary. As shown in Fig.~\ref{fig:globalmotivation}(f), with Top-8 expert selection, cold tokens collectively activate only about 17 experts, while hot tokens span 42. We attribute this disparity to higher routing similarity among cold tokens (Appendix~\ref{why_cold}).

Motivated by this, we propose candidate-restricted routing that applies filtering exclusively to cold tokens, preserving routing fidelity while reducing computation. 
Specifically, for a cold token set $\mathcal{C}_t$ at step $t$, we compute the centroid $\overline{\mathbf{h}} = \frac{1}{|\mathcal{C}_t|} \sum_{i\in\mathcal{C}_t} \mathbf{h}_{\mathrm{in}, i}$, and obtain its routing probabilities $\boldsymbol{\pi}(\bar{\mathbf{h}}) \in \mathbb{R}^E$. The candidate expert set is then defined as $\mathcal{Q} = \operatorname{Top}_{M}\!\left( \boldsymbol{\pi}(\bar{\mathbf{h}}) \right)$, where $K < M < E$. Each cold token $i$ performs Top-$K$ selection strictly within $\mathcal{Q}$:
\begin{equation}
\mathcal{E}_i = \operatorname{Top}_{K}\!\Big( \{ \pi_j(\mathbf{h}_{\mathrm{in},i}) : j \in \mathcal{Q} \} \Big),
\label{eq:cold_restricted_routing}
\end{equation}
where $\pi_j(\mathbf{h}_{\mathrm{in},i})$ denotes the routing probability of token $i$ for expert $j$.
As reported in Table~\ref{tab:candidate-recall}, with $M=48$, the candidate set achieves a Top-8 recall of 97.95\% for cold tokens, compared to 88.34\% for hot tokens, confirming that cold tokens are more amenable to candidate-restricted routing. By limiting cold token expert selection to a set of size $M$, the routing computation cost is reduced from $\mathcal{O}(E)$ to $\mathcal{O}(M)$ per cold token, alleviating the expert-selection overhead. In our experiments, we found that this candidate-restricted routing achieves up to a 13.0\% speedup.

%% file: paper/exp.tex
\section{Experiments}
\label{sec:experiments}

\subsection{Experimental Setup}
\label{sec:experimental_setup}


\textbf{Baselines and configurations.} 
We compare ITC-MoE with D$^2$-MoE~\citep{gu2025d2moe} and TD-MoE~\citep{xu2026tdmoe}, with the uncompressed \textit{Original} model serving as the performance baseline. Following the settings of the compared methods, ITC-MoE compresses only the Gate, Up, and Down projections of the MoE layers. We further compare with TEAM~\citep{team2026}, which focuses on model acceleration without model compression. For a fair comparison, all experiments are conducted on a single NVIDIA A800 GPU, with the same decoding configuration, prompts, and evaluation protocol. The detailed hyperparameters are provided in Appendix~\ref{app:backbone-Model-Details}. Following prior work~\citep{xu2026tdmoe}, we report the compression ratio over the MoE layers only.

\textbf{Models, datasets, and metrics.} 
We conduct zero-shot evaluation on SDAR-30B-A3B-Chat-b32~\citep{sdar} and LLaDA2.0-mini~\citep{bie2025llada20scalingdiffusionlanguage}, covering MBPP~\citep{mbpp}, GSM8K~\citep{gsm}, MultiArith, SingleOp, and SingleQ~\citep{koncel-kedziorski-etal-2016-mawps}. Metrics include task accuracy and end-to-end speedup.

\subsection{Main Results}

\label{sec:main_results}

\textbf{Comparison under Different Compression Ratios.}
Table~\ref{tab:three_methods_two_ratios} compares ITC-MoE with D$^2$-MoE and TD-MoE under different compression ratios on five downstream benchmarks. ITC-MoE consistently achieves substantially higher speedups across the evaluated tasks while maintaining competitive task performance.
In particular, since D$^2$-MoE and TD-MoE are not designed for MoE DLMs, they generally yield limited or even no speedup. For example, at a 30\% compression ratio, D$^2$-MoE only achieves a 0.80$\times$ speedup on MBPP. Moreover, D$^2$-MoE and TD-MoE generally show unstable performance across benchmarks. For example, on MBPP at a 30\% compression ratio, they only achieve scores of 34.24 and 12.45, respectively. In contrast, the proposed ITC-MoE present a score of 47.47 and also maintains stable and competitive performance across different tasks.



\begin{table*}[t]
\centering
\caption{Comparison of D$^2$-MoE~\citep{gu2025d2moe}, TD-MoE~\citep{xu2026tdmoe}, and our ITC-MoE on SDAR-30B-A3B-Chat-b32~\cite{sdar} under different compression ratios. The orange-signed numbers indicate that the corresponding speedups are below $1.00\times$.}
\label{tab:three_methods_two_ratios}
\vspace{-2mm}
\small
\begin{tabular}{cccc|cc|cc}
\toprule
\multirow{2}{*}{Benchmark} & \multirow{2}{*}{Method} & \multicolumn{2}{c|}{10\% Ratio}
& \multicolumn{2}{c|}{20\% Ratio}
& \multicolumn{2}{c}{30\% Ratio} \\
\cmidrule(lr){3-4}
\cmidrule(lr){5-6}
\cmidrule(lr){7-8}
  & 
& Score $\uparrow$ & Speedup $\uparrow$
& Score $\uparrow$ & Speedup $\uparrow$
& Score $\uparrow$ & Speedup $\uparrow$ \\
\midrule

\multirow{4}{*}{MBPP}
& Original
& 65.76 & $1.00\times$
& 65.76 & $1.00\times$
& 65.76 & $1.00\times$ \\
& D$^2$-MoE
& 56.42 & $1.26\times$
& \textbf{52.53} & $1.14\times$
& 34.24 & {\color{orange}$0.80\times$} \\
& TD-MoE
& 58.75 & $1.34\times$
& 51.75 & $1.31\times$
& 12.45 & $1.35\times$ \\
& \textbf{ITC-MoE}
& \textbf{59.92} & $\mathbf{5.68\times}$
& 50.58 & $\mathbf{5.68\times}$
& \textbf{47.47} & $\mathbf{6.61\times}$ \\

\midrule

\multirow{4}{*}{GSM8K}
& Original
& 90.60 & $1.00\times$
& 90.60 & $1.00\times$
& 90.60 & $1.00\times$ \\
& D$^2$-MoE
& 83.47 & {\color{orange}$0.86\times$}
& 77.10 & {\color{orange}$0.89\times$}
& \textbf{66.57} & {\color{orange}$0.82\times$} \\
& TD-MoE
& \textbf{88.48} & $1.10\times$
& 72.18 & {\color{orange}$0.96\times$}
& 39.65 & {\color{orange}$0.93\times$ }\\
& \textbf{ITC-MoE}
& 87.87 & $\mathbf{4.60\times}$
& \textbf{79.38} & $\mathbf{4.37\times}$
& 63.46 & $\mathbf{4.71\times}$ \\

\midrule

\multirow{4}{*}{MultiArith}
& Original
& 98.67 & $1.00\times$
& 98.67 & $1.00\times$
& 98.67 & $1.00\times$ \\
& D$^2$-MoE
& 97.67 & $1.06\times$
& 96.83 & $0.87\times$
& \textbf{98.00} & {\color{orange}$0.88\times$} \\
& TD-MoE
& 98.17 & $1.22\times$
& 96.67 & $1.20\times$
& 73.83 & $1.09\times$ \\
& \textbf{ITC-MoE}
& \textbf{99.17} & $\mathbf{5.27\times}$
& \textbf{99.33} & $\mathbf{5.86\times}$
& 96.33 & $\mathbf{7.22\times}$ \\

\midrule

\multirow{4}{*}{SingleOp}
& Original
& 97.15 & $1.00\times$
& 97.15 & $1.00\times$
& 97.15 & $1.00\times$ \\
& D$^2$-MoE
& \textbf{95.55} & {\color{orange}$0.95\times$}
& \textbf{96.62} & {\color{orange}$0.82\times$}
& \textbf{90.70} & {\color{orange}$0.80\times$} \\
& TD-MoE
& 95.02 & $1.13\times$
& 89.15 & $1.05\times$
& 74.02 & $0.71\times$ \\
& \textbf{ITC-MoE}
& 95.37 & $\mathbf{5.34\times}$
& 95.02 & $\mathbf{5.26\times}$
& 88.08 & $\mathbf{5.56\times}$ \\

\midrule

\multirow{4}{*}{SingleQ}
& Original
& 98.17 & $1.00\times$
& 98.17 & $1.00\times$
& 98.17 & $1.00\times$ \\
& D$^2$-MoE
& 94.50 & {\color{orange}$0.87\times$}
& \textbf{96.33} & {\color{orange}$0.84\times$}
& \textbf{88.83} & {\color{orange}$0.90\times$} \\
& TD-MoE
& 96.33 & $1.20\times$
& 86.24 & $1.18\times$
& 61.47 & $1.00\times$ \\
& \textbf{ITC-MoE}
& \textbf{97.25} & $\mathbf{5.36\times}$
& 95.41 & $\mathbf{5.69\times}$
& 84.40 & $\mathbf{6.16\times}$ \\
\bottomrule
\end{tabular}
\vspace{-3mm}
\end{table*}

\textbf{Comparison with Non-Compression Baseline.}
Table~\ref{tab:cross_backbone_five_benchmarks} compares ITC-MoE with TEAM on SDAR and LLaDA2.0-mini. Note that TEAM improves inference efficiency without compressing model parameters. Therefore, we set the compression ratio of ITC-MoE to a small value of 3\% for a fair comparison.
Across five benchmarks, ITC-MoE achieves substantially higher speedups than TEAM on both models while maintaining comparable task performance. For example, on SDAR, ITC-MoE achieves \(3.46\times\)-\(4.61\times\) speedups versus \(1.74\times\)-\(2.08\times\) for TEAM, while obtaining the highest score on MultiArith and matching the Original model on SingleQ. On LLaDA2.0-mini, it achieves \(2.66\times\)-\(4.53\times\) speedups versus \(1.26\times\)-\(1.72\times\) for TEAM, while achieving the highest score on MBPP and matching the Original model on SingleOp and SingleQ. These results demonstrate that ITC-MoE achieves a more favorable efficiency-performance trade-off than TEAM.


\begin{table*}[t]
\centering
\small
\caption{Comparison of ITC-MoE and TEAM across different backbone models, with model compression applied to ITC-MoE.}
\label{tab:cross_backbone_five_benchmarks}
\vspace{-2mm}   
\begin{tabular}{cccc|cc}
\toprule
\multirow{2}{*}{Benchmark} &
\multirow{2}{*}{Method} &
\multicolumn{2}{c|}{SDAR} &
\multicolumn{2}{c}{LLaDA2.0-mini} \\
\cmidrule(lr){3-4}\cmidrule(lr){5-6} 
&& Score $\uparrow$ & Speedup $\uparrow$
& Score $\uparrow$ & Speedup $\uparrow$ \\
\midrule
\multirow{3}{*}{MBPP}
& Original & 65.76 & $1.00\times$ & 80.54 & $1.00\times$ \\
& TEAM    & \textbf{65.76} & $2.08\times$ & 80.16 & $1.56\times$ \\
& \textbf{ITC-MoE} & 65.57 & $\mathbf{4.61\times}$ & \textbf{80.93} & $\mathbf{4.53\times}$ \\
\midrule
\multirow{3}{*}{GSM8K}
& Original & 90.60 & $1.00\times$ & 91.89 & $1.00\times$ \\
& TEAM    & \textbf{90.30} & $1.83\times$ & \textbf{92.80} & $1.72\times$ \\
& \textbf{ITC-MoE} & 90.08 & $\mathbf{3.46\times}$ & 92.56 & $\mathbf{2.94\times}$ \\
\midrule
\multirow{3}{*}{MultiArith}
& Original & 98.67 & $1.00\times$ & 98.33 & $1.00\times$ \\
& TEAM    & 98.83 & $1.77\times$ & \textbf{98.17} & $1.49\times$ \\
& \textbf{ITC-MoE} & \textbf{99.00} & $\mathbf{4.09\times}$ & \textbf{98.17} & $\mathbf{2.69\times}$ \\
\midrule
\multirow{3}{*}{SingleOp}
& Original & 97.15 & $1.00\times$ & 97.86 & $1.00\times$ \\
& TEAM    & 95.20 & $1.89\times$ & 97.51 & $1.26\times$ \\
& \textbf{ITC-MoE} & \textbf{95.55} & $\mathbf{4.31\times}$ & \textbf{97.86} & $\mathbf{2.66\times}$ \\
\midrule
\multirow{3}{*}{SingleQ}
& Original & 98.17 & $1.00\times$ & 99.08 & $1.00\times$ \\
& TEAM    & 95.41 & $1.74\times$ & 98.17 & $1.52\times$ \\
& \textbf{ITC-MoE} & \textbf{98.17} & $\mathbf{3.89\times}$ & \textbf{99.08} & $\mathbf{2.95\times}$ \\
\bottomrule
\end{tabular}
\vspace{-5mm}
\end{table*}

\subsection{Ablation Studies}
\label{sec:ablation_studies}

We use SDAR-30B-A3B-Chat-b32 as the default model for all ablation studies.

\begin{figure}[t]
    \centering
    \begin{subfigure}[b]{0.4\linewidth}
        \centering
        \includegraphics[width=\linewidth]{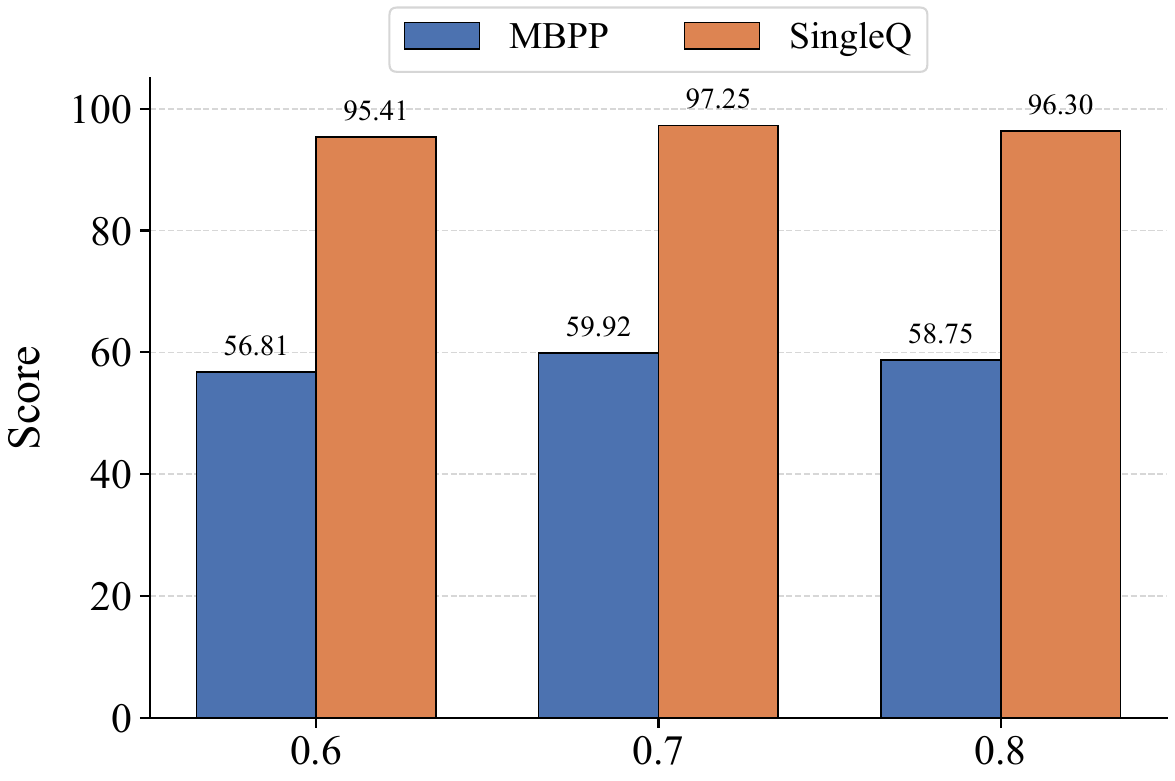}
        \caption{Sensitivity analysis of the hot-token confidence threshold $\tau_h$.}
        \label{fig:ablation_hotcold_threshold}
    \end{subfigure}
    \hspace{5mm}
    \begin{subfigure}[b]{0.4\linewidth}
        \centering
        \includegraphics[width=\linewidth]{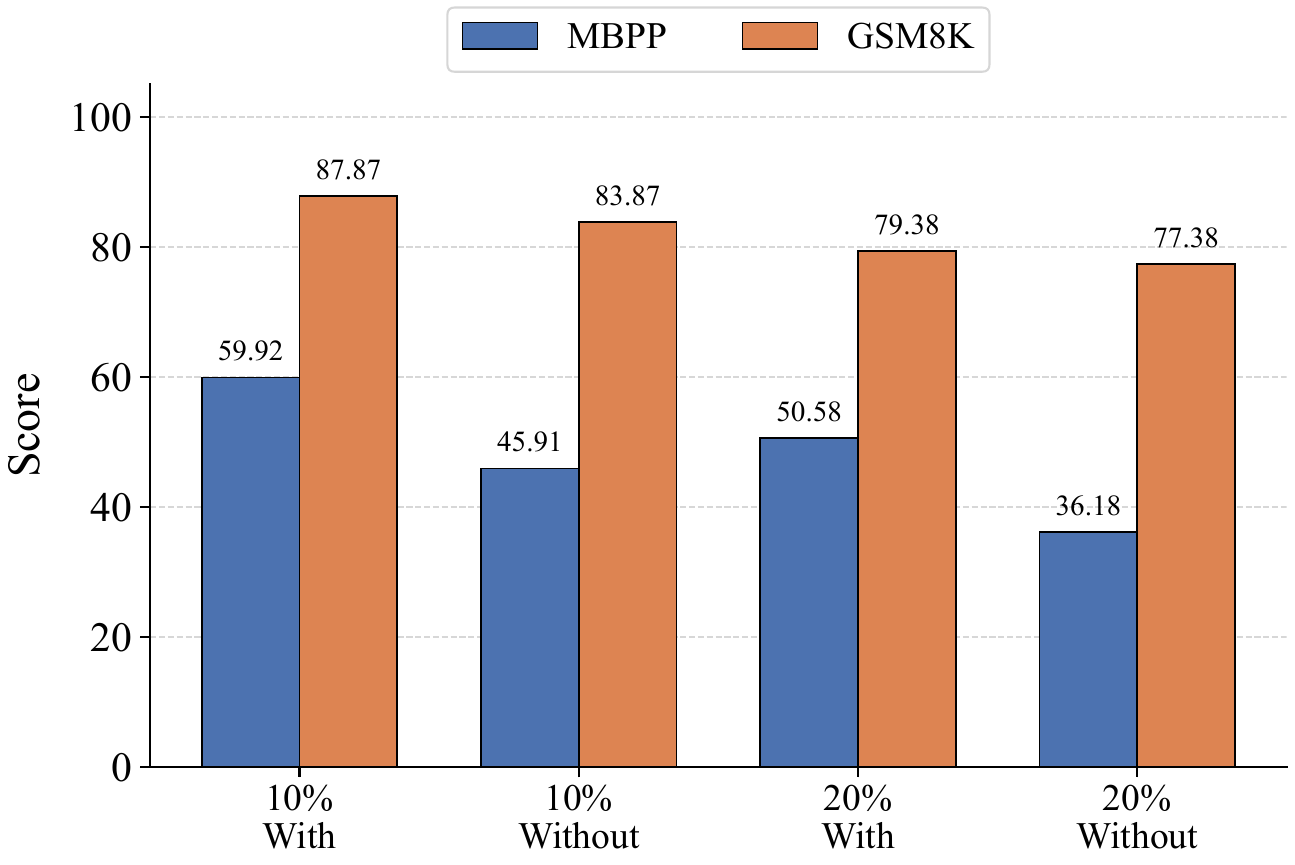}
        \caption{Importance-Weighted Transformation ablation at 10\% and 20\% compression.}
        \label{fig:ablation_whitening}
    \end{subfigure}
    \caption{Hot-token threshold sensitivity and effect of Importance-Weighted Transformation.}
    \label{fig:ablation_combined}
    \vspace{-3mm}
\end{figure}

\textbf{Sensitivity to the Hot-Token Confidence Threshold.}
Fig.~\ref{fig:ablation_hotcold_threshold} examines ITC-MoE's sensitivity to the hot-token confidence threshold $\tau_h$ with $L_h=3$. Varying $\tau_h$ from 0.6 to 0.8 causes only small changes on MBPP and SingleQ, where MBPP ranges from 56.81 to 59.92, and SingleQ remains within 95.41--97.25. The best performance is achieved at $\tau_h=0.7$.

\textbf{Effect of Importance-Weighted Transformation.}
As shown in Figure~\ref{fig:ablation_combined}, the importance-weighted transformation consistently improves performance. At a 10\% compression rate, MBPP increases from 45.91 to 59.92 and GSM8K from 83.87 to 87.87. At 20\%, MBPP increases from 36.18 to 50.58 and GSM8K from 77.38 to 79.38. These results indicate that the importance-weighted transformation effectively mitigates performance degradation caused by compression.

\textbf{Effectiveness of Hot Token Compensation.}
To evaluate the contribution of Hot-token compensation, we reduce the rank of the compensation residual term from 11 to 0 while keeping all other settings unchanged. As shown in Table~\ref{tab:ablation_hot_compensation}, enabling Hot-token compensation improves the MBPP score from 51.75 to 59.92 and the SingleQ score from 94.49 to 97.25, demonstrating its effectiveness in recovering the representation loss introduced by compression.

\textbf{Effectiveness of Learned Rank Allocation.}
As shown in Table~\ref{tab:ablation_fixed_rank}, learned rank truncation consistently outperforms fixed-proportion allocation on both MBPP and GSM8K under the 10\% and 20\% compression ratios. More ablation studies are provided in Appendix~\ref{sec:Additional_ablations}. 

\begin{table*}[t]
\centering

\makebox[\textwidth][c]{%

\begin{minipage}[c]{0.42\textwidth}
\centering
\small
\setlength{\tabcolsep}{3pt}
\captionof{table}{Effect of Hot-token compensation under the 10\% compression ratio.}
\label{tab:ablation_hot_compensation}
\vspace{-2mm}
\begin{tabular}{ccc}
\toprule
\multirow{2}{*}{Setting} & MBPP & SingleQ \\
& Score $\uparrow$ & Score $\uparrow$ \\
\midrule
w.    & \textbf{59.92} & \textbf{97.25} \\
w/o.  & 51.75          & 94.49 \\
\bottomrule
\end{tabular}
\end{minipage}%
\hfill
\begin{minipage}[c]{0.55\textwidth}
\centering
\small
\setlength{\tabcolsep}{3pt}
\captionof{table}{Comparison of learned rank allocation and fixed rank allocation across the three Tucker modes.}
\label{tab:ablation_fixed_rank}
\vspace{-3mm}
\begin{tabular}{cccc}
\toprule
\multirow{2}{*}{Compression Ratio} &
\multirow{2}{*}{Setting} &
MBPP & GSM8K \\
& & Score $\uparrow$ & Score $\uparrow$ \\
\midrule
10\% & Learned ranks      & \textbf{59.92} & \textbf{87.87} \\
10\% & Fixed & 53.09           & 84.00 \\
20\% & Learned ranks      & \textbf{50.58} & \textbf{79.38} \\
20\% & Fixed & 46.19           & 68.00 \\
\bottomrule
\end{tabular}

\end{minipage}
\hfill

}
\vspace{-4mm}
\end{table*}